\documentclass[11pt]{article}
\usepackage[preprint]{acl}
\usepackage{times}
\usepackage{latexsym}
\usepackage[T1]{fontenc}
\usepackage[utf8]{inputenc}
\usepackage{microtype}
\usepackage{inconsolata}
\usepackage{graphicx}
\usepackage{amsmath,amssymb,amsfonts}
\usepackage{booktabs}
\usepackage{booktabs,tabularx,array}
\usepackage{enumitem}
\usepackage{xspace}
\usepackage{amsthm}
\usepackage{subcaption}
\theoremstyle{definition}
\newtheorem{definition}{Definition}[section]
\graphicspath{{figures/}}
\newcommand{\modelname}{MultiEvoSkill\xspace}

\title{\modelname: Structure-Aware Multi-Agent Skill Evolution for Generalizable Skill Artifacts}

\author{Kounianhua Du*$^1$, Jianxing Liu*$^1$, Ruoshui Shan*$^1$, Yucheng Zhang*$^3$,  \\
{\bf Ke Li$^2$, Xing Sun$^2$, Weinan Zhang$^1$}\\
  $^1$Shanghai Jiao Tong University, $^2$ Tencent,  $^3$ Purdue\\
  Shanghai, China\\
  \texttt{\{kounianhuadu, wnzhang\}@sjtu.edu.cn}
  }
  
\begin{document}
\maketitle
\begin{abstract}
Skill documents are becoming the default adaptation surface for
frozen large language model (LLM) agents, but current skill
generators face two key limitations. First, humans and different LLMs differ in how they organize procedural knowledge and approach task
solving. Skills authored from a human or single-model perspective
may therefore fail to align with the reasoning behaviors of other
backbones, limiting cross-model generalization. Second, sequential
skill descriptions inadequately represent procedures with parallel
branches and explicit dependencies, while existing skill
optimizers primarily operate on prose and lack explicit support
for graph-structured skill evolution. We introduce \modelname, a skill-optimization framework comprising three core components: 1) \emph{Skill compiler}
casts the skill as a typed graph with tailored operators, producing a compact, informative representation with explicit procedural topology; 2) \emph{Heterogeneous agent system} proposes revision advices using collective intelligence, where minimal and informative issues are identified based on multiple sets of experiences. 3)  \emph{Structure-aware debate and evolution mechanism} organizes candidate edits into a heterogeneous graph, resolves them under dependency and importance constraints, and distills the collective proposal into the artifact. Multi-dimensional gating ensures the information gain and topological correctness of each evolution. Optimal skills are then obtained when the system converges at a consistent state. Various experiments and ablations against SOTA methods validate the effectiveness and generalization of \modelname.

\end{abstract}

\section{Introduction}
\label{sec:intro}
Large language model (LLM) agents  externalise procedural
knowledge into \emph{skills}: reusable natural-language policies that can be loaded by a frozen backbone at run time.  Anthropic's \textsc{Agent Skills} standard~\citep{anthropic2025skills,
anthropic2025engineering} formalizes this interface, while recent work treats the skill document itself as an optimizable artifact, improving it through execution feedback, trajectory distillation, or evolutionary
search~\citep{skillopt2026,skillgrad2026,trace2skill2026,gepa2025,
mocha2026}. Unlike parameter updates, a skill is lightweight to edit,
easy to inspect, and can in principle be reused by different frozen
models without changing their weights or execution harnesses. This
makes skill evolution a promising route to persistent agent
improvement.

\begin{figure}[t]
\centering

\begin{subfigure}{\columnwidth}
    \centering
    \includegraphics[width=\linewidth]{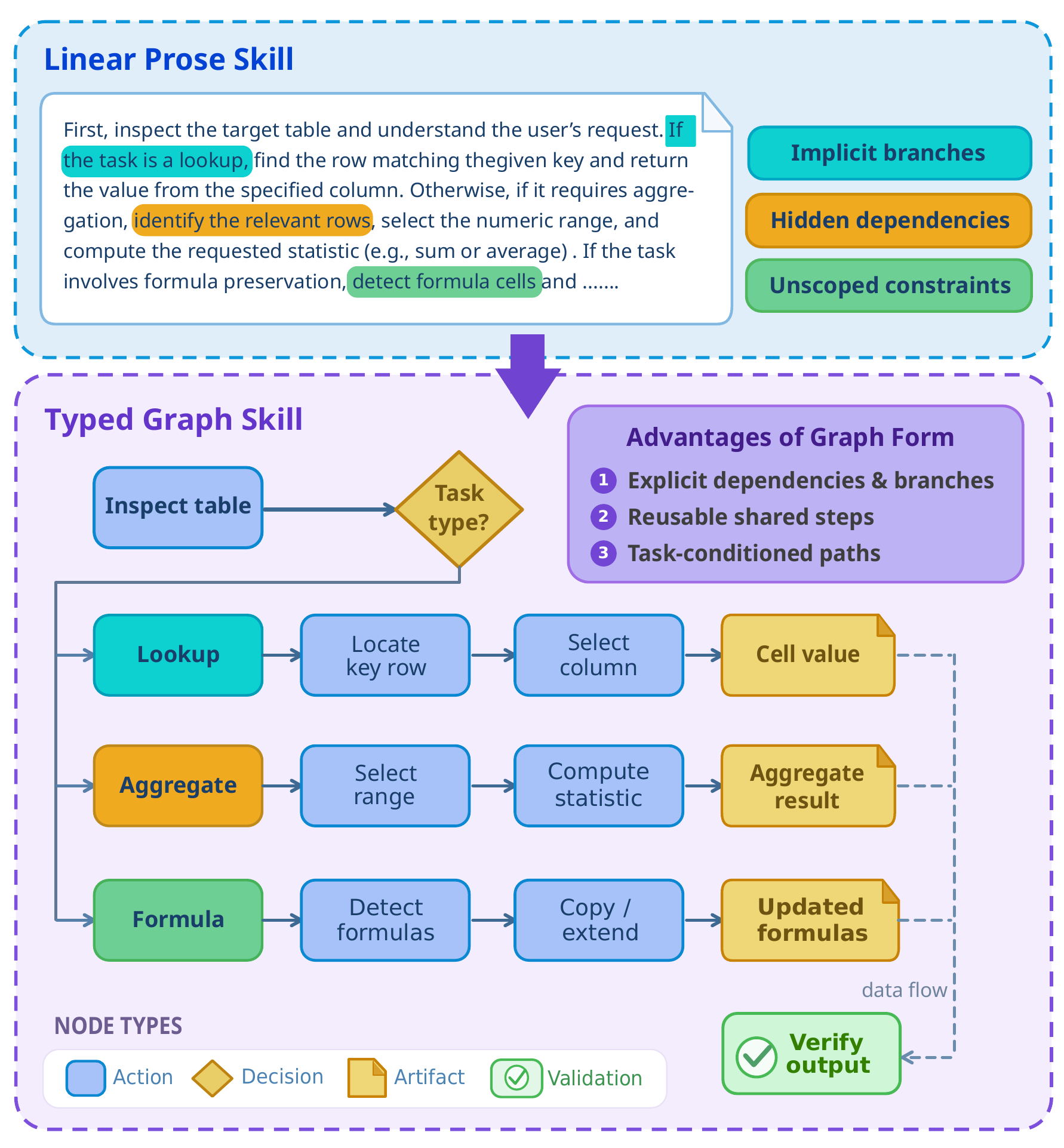}
    \caption{Prose-form vs. graph-form skills.}
    \label{fig:skill_form_illustration}
\end{subfigure}

\vspace{0.5em}

\begin{subfigure}{\columnwidth}
    \centering
    \includegraphics[width=\linewidth]{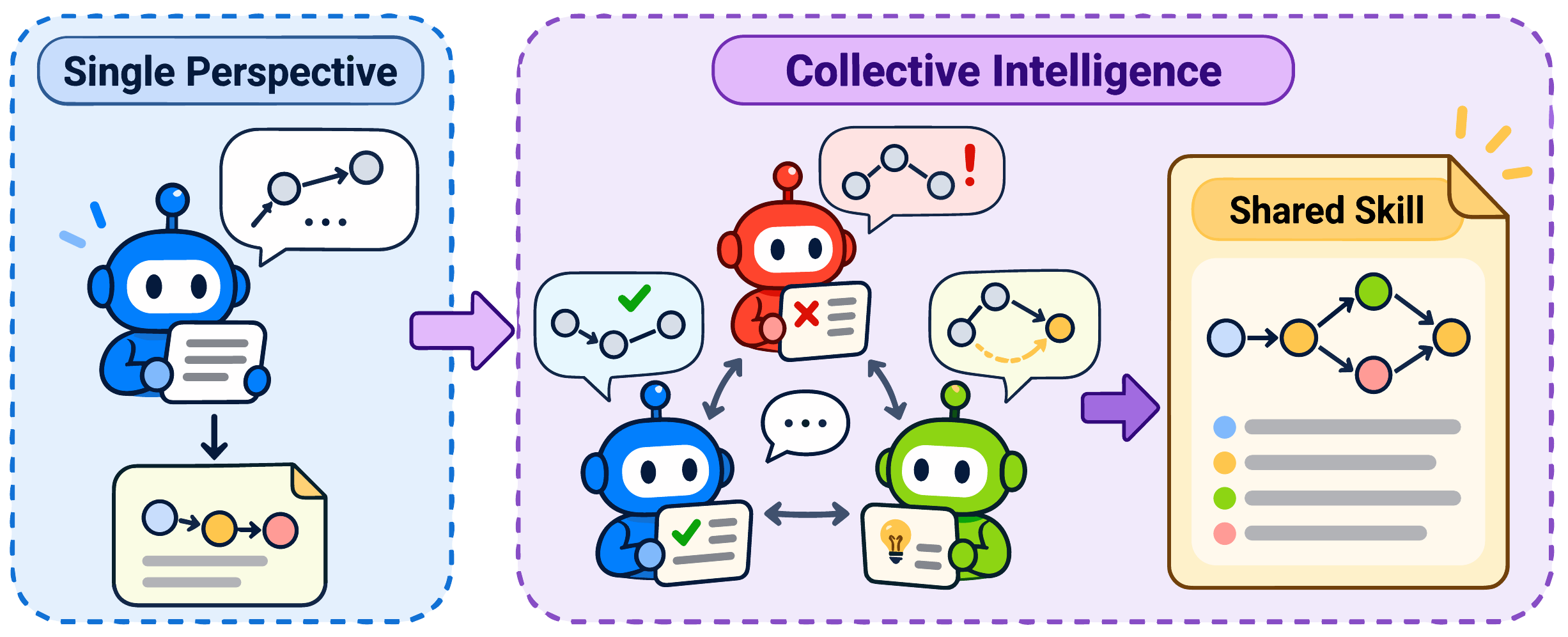}
    \caption{Single-model vs. collective intelligence.}
    \label{fig:collective_intelligence_illustration}
\end{subfigure}

\caption{
Illustration of two key advantages of our approach.
(a) Graph-form skills explicitly encode procedural dependencies,
branch conditions, and reusable actions, unlike linear prose.
(b) Collective intelligence enables heterogeneous agents to
exchange complementary experience and consolidate it into
a single generalizable skill artifact.
}
\label{fig:main_illustration}
\end{figure}

Despite this promise, current skill optimization pipelines face two recurring limitations. First, most methods represent a skill as a linear prose document. Prose is convenient for LLMs to read and write, but its procedural relationships are not exposed as explicit, mechanically addressable objects. In high-reasoning-demand domains, such as mathematical reasoning, spreadsheet manipulation, workflow generation, or tool-grounded question answering, successful task solving often requires explicit dependencies among intermediate artifacts, constraint
checks, conditional decisions, and branch-specific recovery steps. Flattening such structure into a sequential paragraph obscures which steps are prerequisites, which branches are mutually exclusive, and which constraints must be preserved. The limitation is even more evident when cross-sample generalization is required within a single domain: different samples may require different procedural branches, even though they share the same high-level task family. A single prose trajectory
tends to over-specify one path or ambiguously mix several paths, making it difficult to reuse the skill across heterogeneous samples. As illustrated in Figure~\ref{fig:skill_form_illustration}, the graph form makes procedural dependencies and task-conditioned branches explicit, enabling more flexible reuse across heterogeneous samples. Although recent work has explored structured skill libraries, pseudocode skills, and graph-based skill composition~\citep{skillpseudo2026,grasp2026,
skilldag2026,psn2026,asgsi2025}, existing graph-oriented approaches primarily focus on inference-time composition, library organization, or programmatic skill networks. Our focus is the joint design of training-time graph editing, heterogeneous evidence aggregation, and validation-gated updates to one persistent artifact, rather than inference-local composition or library-level organization.

Second, existing skill evolution is often driven by experience from a single backbone or a homogeneous family of models. However, humans and LLMs do not organize procedural knowledge in the same way, and different LLM families exhibit different planning habits, abstraction levels, tool-use strategies, and failure modes. A skill authored by a human or optimized from one model's trajectories may therefore encode model-specific reasoning conventions that do not transfer to other backbones. Simply adding more models is not sufficient: a naive ensemble
that sends every task to every model increases cost and may still fail to distill their complementary evidence into a reusable artifact. What is needed is an evolution mechanism that lets heterogeneous models collect distinct experience, critique and refine each other's diagnoses, and consolidate their collective intelligence into a single skill that remains useful when deployed without an inference-time ensemble. Figure~\ref{fig:collective_intelligence_illustration} contrasts single-model skill evolution with collective evolution, where heterogeneous agents exchange complementary evidence and consolidate it into a shared, reusable skill artifact.

We introduce \modelname, a structure-aware multi-agent framework for evolving generalizable skill artifacts. \modelname represents a skill as a typed attributed directed multigraph whose nodes encode actions, decisions, artifacts, and constraints, and whose edges encode ordering, branching, production, consumption, and semantic dependency relations.
This representation makes the topology of a task-solving procedure explicit. It allows one skill to express complex reasoning structures and to activate different branches for different samples within the same domain.

On top of this representation, \modelname evolves the skill through a heterogeneous multi-agent pipeline. Multiple LLM executors
collect contrasting trajectories for the same training items; reflection agents extract graph-grounded failures and reusable successful behaviors; proposer agents debate and revise bounded graph edits; and a structure-aware revisor composes compatible operations into one candidate update. The result is not a committee at test time, but a single frozen skill graph
that has absorbed the complementary evidence and reasoning patterns of multiple LLMs during training.

\paragraph{Contributions.}
\begin{itemize}[leftmargin=*, itemsep=1pt, topsep=1pt]
    \item We propose the \textbf{typed attributed directed skill graph representation} coupled with \textbf{graph-grounded issue}  and \textbf{typed graph edit operations} that makes procedural topology, typed dependencies, and sample-conditioned branches explicit, improving expressiveness for complex reasoning tasks and cross-sample generalization within a domain.

    \item We develop a \textbf{heterogeneous multi-agent evolution pipeline} where collective intelligence from multiple LLMs is explored to a high extent by multi-LLM execution, debate-driven reflection evolving, and collective experience reuse. 

    \item We propose a \textbf{multi-dimensional gating mechanism} that ensures the information gain of evolution and provides compact feedback for the next loop.
    
    \item \modelname transforms collective intelligence  into generalizable skill artifacts, which works well under complex domain and offer robust cross-backbone performances. 
\end{itemize}

\section{Related Work}
\label{sec:related}

\paragraph{Skill representations and adaptation surfaces.}

File-based natural-language skills externalise procedural knowledge
without modifying model parameters, providing a persistent adaptation
surface formalised by Anthropic's \textsc{Agent Skills} standard
~\citep{anthropic2025skills,anthropic2025engineering} and adopted by
systems that accumulate reusable experience in skill libraries
~\citep{wang2023voyager,zhou2026mementoskills}. More explicit
representations compile prose into typed programs or task-level
graphs: Skill-as-Pseudocode introduces typed contracts
~\citep{skillpseudo2026}, while GraSP constructs and locally repairs
inference-time DAGs~\citep{grasp2026}. At a broader level, SkillDAG,
PSN, and ASG-SI organise skills into persistent graphs or programmatic
networks~\citep{skilldag2026,psn2026,asgsi2025}.

\paragraph{Skill evolution mechanisms.}

Skill evolution continuously improves persistent, reusable skill
artefacts by incorporating execution experience and retaining
validated revisions for future reuse
~\citep{ding2026agentskillevolution,zhou2026agentskillsurvey}. Trajectory-distillation methods compare multiple complete
trajectories to isolate reusable sequence-level patterns from
run-specific noise. Execution-feedback methods instead revise the
current skill directly from step-level signals within individual
runs, such as errors, unmet requirements, or rewards
~\citep{ding2026agentskillevolution}.
Multi-agent systems divide this process among executors,
reflection analysts, proposers, and verifiers,
while population-based methods explore multiple candidates
through mutation, recombination, or fitness-guided selection.
Reinforcement-learning methods optimise skill generation, selection,
or refinement policies using task- or reuse-level rewards accumulated
across rollouts.

Across these routes, text-space updates commonly employ bounded insertion,
replacement, deletion, or targeted rewriting; structured updates
modify the corresponding nodes and relations. Candidate acceptance is
typically controlled by empirical improvement or verifier-based
validity, while evolution terminates upon verifier satisfaction, lack
of further gain, budget exhaustion, or population stability.

\paragraph{Multi-agent methods for skill evolution.}

Individual LLMs are limited in exploring diverse skill updates,
motivating multi-agent methods that exchange information and refine
candidate skills collectively.
Within this framework, proposers generate candidate answers, plans,
skills, or edits; reviewers provide critiques;
verifiers assess candidates using execution evidence, rules, or scores;
revisers modify candidates following critique or evaluation;
executors run skills or workflows to produce trajectories, rewards,
and errors; and specialist agents contribute distinct domain or
task-specific capabilities
~\citep{coevoskills2026,skillgen2026,mindskill2026}.

These roles define functional responsibilities, whereas multi-agent
debate (MAD) provides an important interaction paradigm in which peer
agents iteratively critique and revise competing proposals
~\citep{du2023mad,liang2023encouraging,chan2023chateval,
khan2024debate,madstrategiessurvey2026}. Effective debate benefits
from diverse perspectives and calibrated evidence
~\citep{demystifyingmad2026}, and commonly terminates through
consensus or stability criteria
~\citep{networkedllmconsensus2026,madjudge2025}. Agent contributions
are subsequently reconciled through response fusion, patch
consolidation, or candidate selection
~\citep{jiang2023llmblender,wang2024moa,trace2skill2026,
skillcat2026,agentpso2026}.


\section{Methodology}
\label{sec:method}

\begin{figure*}[t]
\centering
\includegraphics[width=\textwidth]{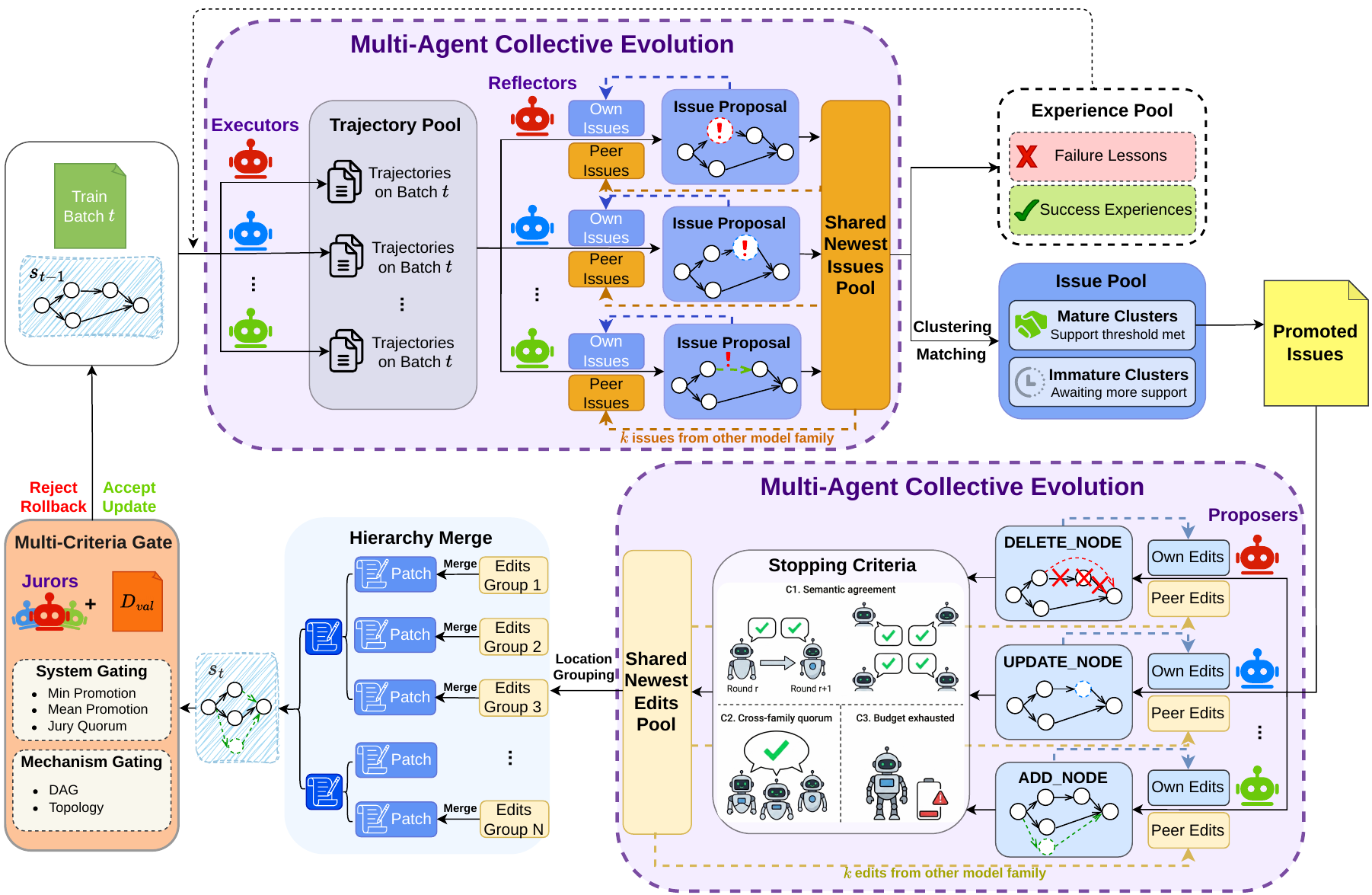}
\caption{Overview of \modelname. Heterogeneous executors collect evidences for reflection; issue-specific proposer debates produce patches against one fixed parent. A hierarchical revisor merges patches, and the gate commits or rejects the resulting graph. Selection data enter the gate, not proposal generation.}
\label{fig:overview}
\end{figure*}

\subsection{Overview}
\modelname evolves one persistent skill graph using heterogeneous execution evidence, graph-grounded reflection, debated edits, hierarchical merging, and validation-gated updates. Let $D_{\mathrm{tr}}$, $D_{\mathrm{sel}}$, and $D_{\mathrm{test}}$ denote disjoint training, selection, and test splits. An initializer constructs $G_0$ from training inputs, or imports a supplied checkpoint. At step $t$, all proposals refer to the same parent $G_{t-1}$ and use a scheduled batch $\mathcal D_t\subseteq D_{\mathrm{tr}}$:
\begin{equation}
\begin{aligned}
\rho_t &= \operatorname{Reflect}(G_{t-1},\mathcal Z_t),\\
\mathcal B_t &= \operatorname{Debate}(G_{t-1},\rho_t),\\
G'_t &= \operatorname{Merge}(G_{t-1},\mathcal B_t),\\
G_{t} &= \operatorname{Gate}(G_{t-1},G'_t).
\end{aligned}
\label{eq:pipeline}
\end{equation}
Here $\mathcal Z_t$ contains trajectories and task scores, $\rho_t$ is the consolidated issue set, and $\mathcal B_t$ is the edit pool. Executor, reflector, proposer, revisor, and juror are functional roles whose model sets may overlap. Figure~\ref{fig:overview} summarizes the loop.

\subsection{Skill Compiler}
The compiler represents procedural knowledge with explicit graph objects rather than leaving all relationships implicit in prose. Natural language still supplies node instructions and edge annotations; graph structure makes these relationships addressable by the optimizer.

\begin{definition}[Typed attributed skill graph]
A skill is a directed multigraph $G=(V,E,\mathcal A,\mathcal R,\psi,\Phi,\alpha)$, where $\psi:V\to\mathcal A$ and $\Phi:E\to\mathcal R$ assign node and edge types, and $\alpha$ associates attributes with graph elements. Nodes represent actions, decisions, artifacts, or constraints; edges represent ordering, branching, production, consumption, or constraint relations.
\end{definition}
Node attributes encode local instructions and applicability conditions. Edge attributes encode guards and dependency explanations. The executor consumes a rendered representation of the skill. In our implementation, a candidate graph is accepted only if all node and
edge references resolve, the graph is acyclic, every executable node is
reachable from START and can reach END, and all node and edge types satisfy
the declared schema.

Table~\ref{tab:graph-document} illustrates the correspondence. A paragraph can describe a subgraph, so this is not an invertible prose-to-graph mapping or a span-level source map.
\begin{table}[t]
\centering
\caption{Graph--document correspondence for the spreadsheet formula/value branch. Original node IDs are retained.}
\label{tab:graph-document}
\small
\begin{tabularx}{\columnwidth}{@{}p{0.28\columnwidth}X@{}}
\toprule
Graph element & Procedural meaning \\
\midrule
Decision 7 & Decide whether results must remain formula-driven. \\
$7\xrightarrow{\mathrm{true}}8$ & Create formulas with correct references when recalculation is required. \\
$7\xrightarrow{\mathrm{false}}9$ & Compute and write materialized values when requested. \\
$8\to10$, $9\to10$ & Continue to row merging after the selected branch. \\
Node 10 label & Preserve existing order; append only missing keys. \\
Edge note & Explain a guard or ordering rationale. \\
\bottomrule
\end{tabularx}
\end{table}

\begin{definition}[Graph-grounded issue]
An issue is an evidence-backed diagnosis tied to a specific graph version:
\begin{equation}
\begin{split}
\rho_{t,k}&=(h_t,\Lambda_k,d_k,u_k,\mathcal E_k,w_k,\xi_k),\\
h_t&=\operatorname{Hash}(G_{t-1}),\qquad \mathcal E_k\ne\varnothing.
\end{split}
\label{eq:issue}
\end{equation}
\end{definition}
Here $\Lambda_k$ identifies nodes, edges, a subgraph, or a global structural gap; $d_k$ describes the deficiency; $u_k$ is a reusable correction rule; and $\mathcal E_k$ contains supporting item IDs. The values $w_k$ and $\xi_k$ estimate impact and regression risk. Non-global references must resolve in $G_{t-1}$. A preserve record $\sigma_k=(b_k,a_k,\mathcal E_k)$ analogously records a successful behavior and its applicability condition. Such evidence supports a diagnosis, not a measured repair benefit.

\begin{definition}[Typed graph edit]
An edit $b_i=(h_t,k_i,O_i,\eta_i)$ targets one issue $k_i$ in parent version $h_t$. It contains an ordered operation list $O_i$ and metadata $\eta_i$ recording rationale, evidence, expected effects, risk, and confidence. Supported operations are \texttt{add\_node}, \texttt{update\_node}, \texttt{delete\_node}, \texttt{add\_edge}, and \texttt{delete\_edge}.
\end{definition}
An edit may contain several operations needed for a local repair. Mechanical applicability and behavioral effectiveness remain separate: successfully applying a patch does not establish that it fixes its target issue.

\subsection{Heterogeneous Evidence and Edit Generation}
\paragraph{Execution.}
The current method description uses multiple executors on the same training item. With $K_{\mathrm{exec}}$ executor models, each $x\in \mathcal D_t$ yields
\begin{equation}
\tau_{t,x}^{(j)}=M_j(G_{t-1},x),\quad j=1,\ldots,K_{\mathrm{exec}}.
\end{equation}
This supplies contrasting model-specific behaviors, but costs more than assigning each item to one executor. It should not be interpreted as disjoint-shard execution.

\paragraph{Reflection and issue debate.}
For each item, a reflector $R_a$ examines the trajectory group $\mathcal T_{t,x}=\{\tau_{t,x}^{(j)}\}_j$ and produces a small set of graph-grounded issues. Reflection compares successful and failed trajectories to identify potentially reusable corrections. A shared failure is evidence of a recurring problem, not proof that the skill rather than shared model limitations caused it. At debate round $r$, reflectors revise their diagnoses using peer feedback:
\begin{equation}
\begin{aligned}
\rho_{t,x}^{(a,0)}&=R_a(G_{t-1},x,\mathcal T_{t,x}),\\
\rho_{t,x}^{(a,r)}&=R_a\bigl(G_{t-1},x,\mathcal T_{t,x},\rho_{t,x}^{(a,r-1)},\\
&\hspace{35mm}\rho_{t,x}^{(\mathcal N_a,r-1)}\bigr).
\end{aligned}
\end{equation}
Here $\mathcal N_a$ denotes the peers shown to reflector $a$.

Final diagnoses are consolidated across the batch. Pending issue clusters accumulate support across batches. Let $\phi(i)$ embed an issue's diagnosis, failure lesson, and proposed rule. An issue joins the most similar representative above threshold $\delta$:
\begin{equation}
C^*(i)=\arg\max_{C:\,s(i,r_C)\ge\delta}s(i,r_C),
\end{equation}
where $s(i,r_C)=\cos(\phi(i),\phi(r_C))$. If no cluster qualifies, a new one is created. The representative $r_C$ is chosen by average within-cluster similarity. Let $\mathcal P_C$ contain the distinct reflector--sample pairs contributing issues to $C$. Cluster support is
\begin{equation}
u(C)=|\mathcal P_C|.
\end{equation}
Clusters with $u(C)\ge m$ become eligible for proposal generation and leave the pending pool. Repeated diagnoses from the same reflector on the same item count once. This threshold need not imply support from $m$ distinct samples. Cross-batch graph references must still resolve against the current parent.

\paragraph{Issue-specific proposal debate.}
Let $P_i$ denote proposer $i$, $\mathcal D_{t,k}$ the evidence-linked examples for issue $\rho_{t,k}$, and $H_t$ the retrieved edit history. Initial edits are
\begin{equation}
O_{t,k,i}^{(0)}=P_i(G_{t-1},\rho_{t,k},\mathcal D_{t,k},H_t).
\end{equation}
Each issue is debated independently against the same parent. At later rounds, a proposer receives its own edit and a sparse neighborhood selected preferentially from other model families:
\begin{equation}
\begin{aligned}
O_{t,k,i}^{(r)}=P_i\bigl(&G_{t-1},\rho_{t,k},H_t,O_{t,k,i}^{(r-1)},\\
&\{O_{t,k,j}^{(r-1)}:j\in\mathcal N_i\}\bigr).
\end{aligned}
\end{equation}
Semantic agreement is measured on rendered candidate skills using embeddings and cosine-based clustering. Debate stops on a cross-family cluster quorum, sufficiently high and stable agreement, or the round budget. Agreement controls communication, not acceptance. The final issue-specific edits form $\mathcal B_t$.

\subsection{Structure-Aware Hierarchical Merging}
\paragraph{Grouping and consolidation.}
The revisor checks each edit against $G_{t-1}$, excludes structurally invalid candidates, and deduplicates equivalent operation sequences. Remaining edits are grouped into bounded minibatches, prioritizing overlapping targets and nearby graph locations. At hierarchy level $\ell$, minibatch $\mathcal B_{t,q}^{(\ell)}$ is consolidated as
\begin{equation}
\begin{aligned}
\delta_{t,q}^{(\ell+1)}=M_{\mathrm{rev}}\bigl(&G_{t-1},\\
&\{(\delta_{t,j}^{(\ell)},\rho_{t,j}):j\in\mathcal B_{t,q}^{(\ell)}\}\bigr),
\end{aligned}
\end{equation}
where $\delta_{t,q}^{(\ell)}$ represents a patch relative to $G_{t-1}$. At higher levels, $\rho_{t,j}$ denotes the evidence and issue provenance carried by the corresponding merged patch. The revisor combines compatible corrections and resolves conflicts. Minimal rewriting may express complementary input corrections, but should not introduce unrelated goals. This is not closed-set selection of immutable operation IDs. All intermediate patches remain relative to the same $G_{t-1}$.

\paragraph{Composition and validation.}
The final operation sequence $\delta_t$ respects operation-application dependencies, such as creating a node before adding an incident edge. These differ from task-execution dependencies represented in the skill. Let $\widetilde G_t=G_{t-1}\oplus\delta_t$, where $\oplus$ applies operations in order and yields $\bot$ on failure. Then
\begin{equation}
G'_t=\begin{cases}
\widetilde G_t,&\delta_t\ne\varnothing,\ \widetilde G_t\ne\bot,\\
&V_{\mathrm{struct}}(\widetilde G_t)=1,\\
G_{t-1},&\text{otherwise.}
\end{cases}
\end{equation}
Only the realized composition receives behavioral validation; constituent edits do not acquire individual causal credit from a successful merge.

\subsection{Multi-Dimensional Gating and Evolution}
\paragraph{Paired behavioral evaluation.}
A fixed assignment $\{(D_{\mathrm{sel}}^j,E_j)\}_{j=1}^{K_{\mathrm{val}}}$ pairs each juror with selection examples before evolution. Parent and candidate are evaluated using the same assignment. For a task-specific score $\mathcal R$, $\mathcal R(E_j,G;D_{\mathrm{sel}}^j)$ is defined as the score of skill $G$ executed by juror $E_j$ on selection set $D_{\mathrm{sel}}^j$.
\begin{equation}
\begin{aligned}
\Delta_j={}&\mathcal R(E_j,G'_t;D_{\mathrm{sel}}^j)\\
&-\mathcal R(E_j,G_{t-1};D_{\mathrm{sel}}^j).
\end{aligned}
\end{equation}
The gate uses worst-juror change, mean change, and the fraction of jurors with non-decreasing scores:
\begin{equation}
\begin{aligned}
\mathrm{minD}&=\min_j\Delta_j,\\
\mathrm{meanD}&=\frac{1}{K_{\mathrm{val}}}\sum_j\Delta_j,\\
Q&=\frac{|\{j:\Delta_j\ge0\}|}{K_{\mathrm{val}}}.
\end{aligned}
\end{equation}


\paragraph{Structural and semantic checks.}
$V_{\mathrm{struct}}$ checks the graph schema and START-to-END DAG condition in the stated formulation. The topology jury assesses whether dependencies are supported by node meanings. High-confidence hard issues include missing prerequisites, contradictory dependencies, and unsupported dependencies; redundancy or ambiguity alone is a warning. Define $V_{\mathrm{topo}}=1$ when no qualifying hard issue is found. This is an LLM-based review, not a proof of semantic correctness. 

For a graph candidate, acceptance requires
\begin{equation}
\begin{aligned}
A_t={}&[\mathrm{minD}\ge-\epsilon]\wedge[\mathrm{meanD}\ge\tau]\\
&\wedge[Q\ge q_0]\wedge[V_{\mathrm{struct}}=1]\\
&\wedge[V_{\mathrm{topo}}=1].
\end{aligned}
\label{eq:gate}
\end{equation}
The tolerance $\epsilon\ge0$ bounds allowed selection-set regression; $\tau$ specifies required mean gain. The state update is
\begin{equation}
G_{t}=\begin{cases}G'_t,&A_t=1,\\G_{t-1},&A_t=0.\end{cases}
\end{equation}
These five predicates implement three behavioral requirements and two validity checks. 


\section{Experiments}
\label{sec:experiments}

\begin{table*}[!t]
\centering
\caption{Test-set scores across five benchmarks. 
Bold indicates the best result in a column. 
}
\label{tab:main}
\small
\begin{tabular}{@{}lccccc@{}}
\toprule
Method & WikiHow & ToolAlpaca & WebShop & SearchQA & SpreadsheetBench  \\
\midrule
Trace2Skill & 53.04 & 35.30 & 28.69 & 86.88 & 50.00  \\
CoEvoSkills & 53.90 & 33.13 & 31.58 & 87.78 & 83.93  \\
SkillOpt & 53.39 & 53.54 & 34.54 & 88.14 & 82.14  \\
SkillBoost & 54.24 & 35.81 & 28.72 & 80.80 & 83.21  \\
SkillGen & 54.56 & 33.65 & 29.12 & 87.38 & 58.21 \\
\midrule
\modelname & \textbf{55.58} & \textbf{53.88} & \textbf{34.77} & \textbf{88.25} & \textbf{86.07}  \\
\bottomrule
\end{tabular}
\end{table*}

\subsection{Research Questions}
We conduct empirical studies based on the research questions below:
\begin{itemize}
    \item \textbf{(RQ1)} Can \modelname outperform the sota baselines?
    
    \item \textbf{(RQ2)} How does our graph form skill offers robustness?
    \item \textbf{(RQ3)} Does the heterogeneous intelligence brings performance gains? Are there evolved intelligence from the model families?
    \item \textbf{(RQ4)} Does each of the gating mechanism contribute to the final performance?
\end{itemize}

\subsection{Setup}
\label{sec:setup}
\paragraph{Benchmarks and metrics.}
Table~\ref{tab:main} contains WikiHow, ToolAlpaca, WebShop, SearchQA~\citep{dunn2017searchqa} and SpreadsheetBench~\citep{ma2024spreadsheetbench}. The first three datasets are workflow-planning adapters reporting mean per-instance node F1, rather than interactive environment success; WikiHow uses the WorFBench domain~\citep{qiao2024benchmarking}, and SpreadsheetBench uses hard instance accuracy. SearchQA is a context-based question-answering benchmark, using macro-averaged token F1 as its metric. Declared splits and remaining metric qualifications are given in Appendix~\ref{app:protocol}.

\paragraph{Baselines and model configuration.}
The comparison includes feedback-driven SkillOpt~\citep{skillopt2026} and SkillBoost~\citep{skillboost2026}, trajectory-based Trace2Skill~\citep{trace2skill2026}, and multi-agent methods SkillGen~\citep{skillgen2026} and CoEvoSkills~\citep{coevoskills2026}. All methods use a common authoring backbone, \textsc{GPT-5.6-Luna}; \modelname additionally uses proposer identifiers \textsc{Gemini-3.8-Flash}, \textsc{MiniMax-M3}, and \textsc{DeepSeek-V4-Flash}. Each method was compared in a fair manner, on the same random seed and data splits, within the scope of budget-equivalent comparisons.

\subsection{Main Results (RQ1)}
\label{sec:main}
\modelname obtains the highest reported score on every column.
These results demonstrate the exceptional generalization capacity of our framework. The superior performance is primarily driven by two core methodological designs: 1) \modelname utilizes topological representation via typed skill graphs instead of traditional linear prose or trajectory-based methods. The explicit topological structure enables executors to strictly follow prerequisite constraints and branch guards, while allowing for localized, graph-grounded edits to occur during the evolution process without introducing non-local side effects or corrupting orthogonal execution branches. 2) \modelname overcomes the biases and blind spots of a single model family through its multi-agent and -role pipeline. By leveraging contrasting execution trajectories and cross-family peer debates, cognitive weaknesses or missing parameter constraints in one model are systematically compensated by the complementary strengths of other models, ultimately crystallizing robust, high-quality skill artifacts into a unified skill graph.

\subsubsection{(RQ2) Case Study on Skill Forms}
\begin{table*}[!t]
\centering
\small
\begin{tabularx}{\textwidth}{@{}p{0.14\textwidth}XXX@{}}
\toprule
\textbf{Sample type} & \textbf{Required procedure} & \textbf{Why prose is fragile} & \textbf{Graph behavior} \\
\midrule
Lookup &
Identify key row and target column; return exact cell value. &
Aggregation and formatting rules are irrelevant but appear in the same prose skill. &
Activate lookup branch only; skip aggregation and formatting subgraphs. \\
\midrule
Aggregation &
Infer row filter, select numeric range, compute requested statistic, validate units. &
Dependency order is implicit; model may compute before fully resolving the range. &
Use explicit prerequisite and data-flow edges from range selection to computation. \\
\midrule
Formula preservation &
Detect formula-bearing cells, determine dependency relation, copy or extend formula safely. &
Global instruction to ``compute values'' can conflict with formula preservation. &
Activate formula branch guarded by formula-detection decision node. \\
\midrule
Conditional formatting &
Parse condition, locate cells, apply formatting, verify affected region. &
Formatting is a separate operation but prose may mix it with value computation. &
Activate formatting branch with its own validation node. \\
\bottomrule
\end{tabularx}
\caption{The illustrative prose skill combines several workflows in one paragraph, whereas a typed skill graph explicitly represents
different branches for different samples while sharing common upstream nodes.}
\label{tab:prose_vs_graph_case}
\end{table*}

Consider a table-manipulation task family (Table~\ref{tab:prose_vs_graph_case}) in which samples share an initial inspection step but require different procedures: exact lookup, aggregation, formula preservation, or conditional formatting. A prose-form skill can describe all four branches, yet the executor must infer their control flow and dependencies from text, potentially mixing irrelevant operations, computing before resolving the target range, or interpreting conditions such as ``preserve formulas when possible'' inconsistently across backbones. Repairs may also have non-local effects: adding ``always preserve formulas'' to address one failure can interfere with tasks that explicitly require formula replacement. A typed skill graph instead represents actions, decisions, intermediate artifacts, constraints, and validation checks as separately addressable nodes, connected by control-flow, data-flow, prerequisite, and branch-guard edges. Shared inspection nodes can lead to distinct branches, with range selection preceding aggregation, formula detection preceding formula modification, and branch-specific checks validating each result; a repair can then target the relevant branch rather than introduce a globally visible instruction. The distinction is not that natural language cannot express these procedures, but that graph structure makes branch conditions and dependencies explicit targets for inspection and editing.

\subsection{Component Ablations}

\subsubsection{(RQ3) Collective Intelligence and Evolution}
\begin{table}[!h]
\centering
\caption{Proposer-panel ablations.}
\vspace{-10pt}
\label{tab:panel-ablation}
\small
\begin{tabular}{@{}lrrr@{}}
\toprule
Variant & WikiHow & ToolAlpaca & WebShop \\
\midrule
w/o $\delta_1$ & 45.88 & 27.71 & 31.53 \\
w/o $\delta_2$ & 50.75 & 43.85 & 32.88 \\
w/o $\delta_3$ & 50.75 & 45.69 & 33.03 \\
w/o $\delta_4$ & 54.64 & 51.69 & 33.45 \\
\midrule
Full & \textbf{55.58} & \textbf{53.88} & \textbf{34.77} \\
\bottomrule
\end{tabular}
\end{table}
Table~\ref{tab:panel-ablation} presents the ablation results regarding the heterogeneous proposer panel, where $\delta_1$ (\textsc{GPT-5.6-Luna}), $\delta_2$ (\textsc{Gemini-3.8-Flash}), $\delta_3$ (\textsc{DeepSeek-V4-Flash}), and $\delta_4$(\textsc{MiniMax-M3}) are individually removed.
Table~\ref{tab:panel-ablation} likewise favors the full proposer panel on all three reported domains. The largest drops are 9.70 points on WikiHow, 26.17 on ToolAlpaca, and 3.24 on WebShop, all in the $\delta_1$ row. The other three variants lose 0.94--10.03 points. ToolAlpaca and WikiHow involve complex API parameter structures or multiple actions and long-horizon logical chains where single models may exhibit systematic blind spots. In contrast, The collaboration across distinct model architectures enables complementary reasoning paths which combine strong macro-planning with fine-grained tool constraint verification, thereby validating the effectiveness of our heterogeneous collective intelligence design.

\subsubsection{(RQ4) Gating Ablations}
\label{sec:ablation}
\begin{table}[!h]
\centering
\caption{Gating ablations. }
\vspace{-10pt}
\label{tab:gate-ablation}
\small
\begin{tabular}{@{}lrrr@{}}
\toprule
Variant & WikiHow & ToolAlpaca & WebShop \\
\midrule
w/o $g_1$ & 42.32 & 45.95 & 30.53 \\
w/o $g_2$ & 41.23 & 45.40 & 30.11 \\
w/o $g_3$ & 40.84 & 47.06 & 31.35 \\
w/o $g_4$ & 49.07 & 46.88 & 33.31 \\
w/o $g_5$ & 53.19 & 49.02 & 33.24 \\
\midrule
Full Gate & \textbf{55.58} & \textbf{53.88} & \textbf{34.77} \\
\bottomrule
\end{tabular}
\end{table}
Table 5 evaluates the contribution of each component within our multi-dimensional gating framework, including behavioral gates ($g_1: \text{minD}$, $g_2: \text{meanD}$, $g_3: Q$) and structural/topological gates ($g_4: V_{\text{struct}}$ and $g_5: V_\text{topo}$)
Our gating components are shown to be essential, as removing any single gate results in performance drops ranging from 1.46 to 14.74 points.
\begin{itemize}[leftmargin=10pt]
    \item Behavioral Gating ($g_1, g_2, g_3$): Acting as the empirical gain and consensus filter, the absence of minimum degradation checks ($g_1$) or jury majority consensus ($g_3$) permits sub-optimal or regressive patches to infiltrate the skill pool, leading to severe performance drops on interactive tasks like ToolAlpaca (dropping to 45.40 without $g_2$) and WikiHow (dropping to 40.84 without $g_3$).
    \item Structural Gating ($g_4, g_5$): Serving as the structural legitimacy firewall, removing $g_4$ impairs the system's ability to filter out malformed graph modifications like dangling nodes or circular dependencies. And without $g_5$ verifying whether the graph edges encode coherent task dependencies, candidate edits with missing prerequisites, contradictory links, or redundant paths may be accepted as positive updates. Removing these gates leads to a prominent drop across workflow-heavy benchmarks.
\end{itemize}

These results confirm that every component of the multi-dimensional gating is vital for guaranteeing monotonic and structurally sound skill evolution.

\section{Conclusion}
\label{sec:conclusion}
We introduce \modelname, a structure-aware multi-agent framework that distills heterogeneous execution experience into a single reusable skill artifact for frozen LLM agents. Its typed graph representation makes procedural branches and dependencies explicit targets for revision, while evidence-grounded reflection, issue-specific debate, and hierarchical patch consolidation integrate complementary corrections. Multi-dimensional gating combines behavioral evaluation with structural and semantic checks to control candidate acceptance. MultiEvoSkill achieves the highest reported scores across the five benchmarks. Leave-one-out ablation validates the effectiveness of the collective intelligence and evolution mechanism. Ablation on gatings justify the effectiveness of each gating configuration. Together, these findings support structured collective skill evolution as a promising approach to persistent agent improvement.
\section*{Limitations}
\label{sec:limitations}
A structural validator checks graph admissibility, not all intended task semantics. An LLM topology jury can miss unsupported dependencies or reject useful edits. Positive gains on a repeatedly reused selection set can overfit that set, and a state-similarity stopping rule does not imply optimality. The graph is a procedural representation rather than a formally verified execution engine.

\section*{AI Use Statement}

Generative AI tools were used to assist with literature search and summarization, code development and debugging, drafting and editing portions of the manuscript. All AI-assisted suggestions, code, and written content were reviewed and, where applicable, independently verified by the authors. The authors made all final decisions regarding the research methodology, experimental design, analysis, and scientific claims, and take full responsibility for the content of this work.

\bibliography{references}
\appendix
\clearpage
\section{Routes for Skill Evolution}
\label{app:taxonomy}
Existing methods can be grouped by
their update signal and search strategy into trajectory distillation,
execution-feedback refinement, multi-agent evolution,
population-based evolution, and reinforcement-learning-based evolution, 
as summarized in Table~\ref{tab:skill-evolution-taxonomy}.

\begin{table}[h]
\centering
\small
\caption{Major routes for agentic skill evolution.}
\label{tab:skill-evolution-taxonomy}
\begin{tabularx}
{\columnwidth}{@{}p{0.35\columnwidth}X@{}p{0.65\columnwidth}X}
\toprule
\textbf{Route} & \textbf{Representative methods} \\
\midrule
Trajectory distillation
& Trace2Skill~\citep{trace2skill2026},
SkillCAT~\citep{skillcat2026},
SkillEvolver~\citep{skillevolver2026},
Memento-Skills~\citep{zhou2026mementoskills},
WikiSkill~\citep{wikiskill2026} \\

Execution-feedback refinement
& SkillOpt~\citep{skillopt2026},
SkillBoost~\citep{skillboost2026},
SkillGrad~\citep{skillgrad2026},
SkillRevise~\citep{skillrevise2026},
EmbodiSkill~\citep{embodiskill2026} \\

Multi-agent evolution
& CoEvoSkills~\citep{coevoskills2026},
SkillGen~\citep{skillgen2026},
MIND-Skill~\citep{mindskill2026},
SkillHone~\citep{skillhone2026} \\

Population-based evolution
& AgentPSO~\citep{agentpso2026},
MOCHA~\citep{mocha2026},
GEPA~\citep{gepa2025},
EvoSkill~\citep{evoskill2026} \\

Reinforcement-learning-based
& D2Skill~\citep{d2skill2026},
SkillRL~\citep{skillrl2026},
SkillOS~\citep{skillos2026},
Skill1~\citep{skill12026} \\
\bottomrule
\end{tabularx}
\end{table}

\section{Running Configuration}
\label{app:protocol}
All experiments use the following fixed configuration in Tables~\ref{tab:hyperparameters} and \ref{tab:splits}. Common declarations are seed 42, at most two additional batch attempts, two reflection rounds, two peer edits per proposal revision, one to three proposal-revision rounds, issue similarity threshold 0.88, debate similarity threshold 0.95, cluster support threshold 2 and hierarchical merge minibatch size 8. The embedding identifier is \texttt{text-embedding-3-small}.


\begin{table}[t]
\centering\small
\caption{Evolution hyperparameters used in our experiments: initialization examples $n_0$, epochs $e$,
batch size $b$, regression base tolerance $\epsilon$, mean-gain threshold $\tau$,
and non-decrease fraction $q_0$.}
\label{tab:hyperparameters}
\begin{tabular}{@{}lrrrrrrr@{}}
\toprule
Domain & $n_0$ & $e$ & $b$ & $\epsilon$ & $\tau$ & $q_0$  \\
\midrule
WikiHow & 40 & 4 & 40 & .03 & .01 & .5  \\
ToolAlpaca & 40 & 4 & 40 & .03 & .01 & .5  \\
WebShop & 40 & 4 & 40 & .03 & .01 & .5  \\
SpreadsheetBench & 40 & 4 & 40 & .10 & .02 & .5  \\
SearchQA & 40 & 4 & 40 & .02 & .01 & .5  \\
\bottomrule
\end{tabular}
\end{table}

\begin{table}[t]
\centering\small
\caption{Dataset split sizes and evaluation protocol.}
\label{tab:splits}
\begin{tabular}{llll}
\toprule
Domain & Train & Selection & Test \\
\midrule
WikiHow & 157 & 53 & 52 \\
ToolAlpaca & 56 & 19 & 18 \\
WebShop & 80 & 27 & 26 \\
SpreadsheetBench & 80 & 40 & 280 \\
SearchQA & 400 & 200 & 1400 \\
\bottomrule
\end{tabular}
\end{table}

\section{Prompt Responsibilities}
\label{app:prompts}
\begin{description}[style=nextline]
\item[Initializer.] Infer reusable procedures from training inputs and emit nodes, types, attributes, and edges. Avoid reference answers and changes to immutable task interfaces.
\item[Reflector.] Diagnose trajectory-supported issues, distinguish reusable lessons from sample-local facts, identify graph locations, and preserve successful behaviors with applicability conditions.
\item[Proposer.] Produce issue-specific bounded edits, record supporting evidence and expected effects, and revise against peer feedback while retaining the same parent graph.
\item[Hierarchical revisor.] Consolidate compatible input corrections, resolve conflicts, and limit rewriting to supported goals. Retain issue provenance and validate the resulting patch.
\item[Topology jury.] Inspect dependency meanings conservatively, distinguish hard issues from warnings, and report confidence and disagreement. A candidate is rejected if any topology jury identifies a high-confidence
hard issue.
\end{description}

\end{document}